\documentclass[a4paper, 10pt, conference]{ieeeconf}      

\usepackage{FG2017}
\usepackage[space]{cite}
\usepackage[hyperpageref]{backref}
\usepackage{graphicx}
\usepackage{amsmath}
\usepackage{amssymb}
\usepackage{algorithm}
\usepackage[]{algpseudocode}

\usepackage{gensymb}
\usepackage{mathptmx}
\usepackage{multirow}
\usepackage[caption=false]{subfig}
\usepackage[flushleft]{threeparttable}
\usepackage{rotating}
\usepackage[export]{adjustbox}
\usepackage{stackengine}
\usepackage[pagebackref=true,breaklinks=true,letterpaper=true,colorlinks,bookmarks=false]{hyperref}

\hypersetup{
backref=true,
bookmarksnumbered,
pdfstartview={FitH},
citecolor={red},
linkcolor={red},
urlcolor={black},
pdfpagemode={UseOutlines}
}

\FGfinalcopy

\IEEEoverridecommandlockouts                              
\title{\LARGE \bf
Spatio-Temporal Facial Expression Recognition Using Convolutional Neural Networks and Conditional Random Fields
}

\author{\parbox{16cm}{\centering
    {\large Behzad Hasani, and Mohammad H. Mahoor }\\
    {\normalsize
    Department of Electrical and Computer Engineering, University of Denver, Denver, CO  \\
    \tt\small{behzad.hasani@du.edu, and mmahoor@du.edu}}}
    \thanks{This work is partially supported by the NSF grants IIS-1111568 and CNS-1427872. }
}

\begin{document}
\IEEEoverridecommandlockouts\pubid{\makebox[\columnwidth]{978-1-5090-4023-0/17/\$31.00~\copyright{}2017 IEEE \hfill} \hspace{\columnsep}\makebox[\columnwidth]{ }}
\ifFGfinal
\thispagestyle{empty}
\pagestyle{empty}
\else
\author{Anonymous FG 2017 submission\\-- DO NOT DISTRIBUTE --\\}
\pagestyle{plain}
\fi
\maketitle

\begin{abstract}

Automated Facial Expression Recognition (FER) has been a challenging task for decades. Many of the existing works use hand-crafted features such as LBP, HOG, LPQ, and Histogram of Optical Flow (HOF) combined with classifiers such as Support Vector Machines for expression recognition. These methods often require rigorous hyperparameter tuning to achieve good results. Recently Deep Neural Networks (DNN) have shown to outperform traditional methods in visual object recognition. In this paper, we propose a two-part network consisting of a DNN-based architecture followed by a Conditional Random Field (CRF) module for facial expression recognition in videos. The first part captures the spatial relation within facial images using convolutional layers followed by three Inception-ResNet modules and two fully-connected layers. To capture the temporal relation between the image frames, we use linear chain CRF  in the second part of our network. We evaluate our proposed network on three publicly available databases, viz. CK+, MMI, and FERA. Experiments are performed in subject-independent and cross-database manners. Our experimental results show that cascading the deep network architecture with the CRF module  considerably increases the recognition of facial expressions in videos and in particular it outperforms the state-of-the-art methods in the cross-database experiments and yields comparable results in the subject-independent experiments.
\end{abstract}

\section{INTRODUCTION}\label{sec:1}

Facial expression is one of the primary non-verbal communication methods for expressing emotions and intentions. Ekman \emph{et al.} identified six facial expressions (viz.  anger, disgust, fear, happiness, sadness, and surprise) as basic emotional expressions that are universal among human beings \cite{c5}. Automatizing the recognition of facial expressions has been a topic of study in the field of computer vision for several years. Automated Facial Expression Recognition (FER) has a wide range of applications such as human-computer interaction, developmental psychology and data-driven animation. Despite the efforts made in developing FER systems, most of the existing approaches either have poor recognition rates suitable for practical applications or lack generalization due to the variations and subtlety of facial expressions \cite{c2}. The FER problem becomes even harder when we recognize expressions in videos. 

Numerous  computer  vision  and  machine  learning  algorithms have been proposed for automated facial expression recognition. Traditional machine learning approaches such as Support Vector Machines (SVM) and Bayesian classifiers perform fine on classifying facial expressions  collected in controlled environments and lab settings. However, these traditional approaches mainly consider still images independently and do not consider the temporal relations of the consecutive frames in a video.
 
 In recent years,  due  to  the  increase  in  the availability of computational power, neural networks methods have become  popular in the research community. In the field of FER, we can find many promising results obtained using Deep Neural Networks \cite{c4,c53}. While in the traditional approaches features were handcrafted, the DNNs  have the ability  to extract more appropriate features from the training images that yield in better visual pattern recognition systems.  Therefore, it has been concluded that the DNNs are able to extract features that generalize well  to  unseen  scenarios and samples. 
 
 Due to uniqueness of expressions for each person and insufficient number of examples in available databases,  one-shot methods have received attentions in recent years~\cite{jiang2016robust,cruz2014one}.  In these methods, information  about different categories are learned from one or very few samples. In this work, since inputs are sequences of frames, the number of training samples is considerably lower than frame-based approaches. Although our method is not purely one-shot, it uses low number of samples in the training phase which makes it is highly generalizable and learns well even with few training samples (Table~\ref{accuracy_subj_ind}).

 In this paper, we look at the problem of facial expression recognition as a two-step learning process.  In the first step, we model the spatial relations within the images. In the second step, we model the temporal relations between consecutive frames in a video sequence and try to predict the labels of each frame while considering the labels of adjacent frames in the sequence. We apply our experiments on three facial expression datasets in order to recognize the six basic expressions along with a neutral state. All of these databases have few number of training samples especially for a DNN. Furthermore, we examine the ability of our model in cross-database classification tasks. 
 
Residual connections introduced by He \emph{et al.} have shown remarkable improvement in recognition rates ~\cite{c6}.  Inspired by ~\cite{c7}, we propose a residual neural network for the first part of our network. For the second part, we use linear chain Conditional Random Fields (CRFs) model \cite{c8}. Cascading of the aforementioned methods is a good approach in modeling facial expressions since it will extract both the spatial and temporal relations of the images in the sequence. 

The remainder of the paper is organized as follows: Section \ref{sec:2}  provides an overview of the related work in this field. Section \ref{sec:3} explains the network proposed in this research. Experimental results and their analysis are presented in Sections \ref{sec:4} and \ref{sec:5}. The paper is concluded in Section \ref{sec:6}. 
\section{RELATED WORK}\label{sec:2}

Recent approaches in the field of visual object recognition, and automated facial expression recognition have used Deep Neural Networks.  These networks  use the huge amount of computing power provided by GPUs, and provide a learning approach  which  can  learn  multiple  levels  of  representation  and  abstraction which allow algorithms to discover complex patterns in the data.

Szegedy \emph{et al.} \cite{c18} introduced GoogLeNet architecture which uses a novel multi-scale approach by using multiple classifier structures, combined with multiple sources for back propagation. Their architecture increases both  width  and  depth of the network  without causing significant penalties. The architecture is composed of  multiple  ``Inception"  layers, which applies convolution on the input feature map  in different scales, allowing the architecture to make more complex decisions. Different variations of the Inception layer have been proposed \cite{c19,c20}. These architectures have shown remarkable recognition results in object recognition tasks.  Mollahosseini \emph{et al.} \cite{c4} have used the traditional Inception layer for the task of facial expression recognition and achieved state-of-the-art results.

Residual connections were introduced by He \emph{et al.}  \cite{c6}. ResNets consist of many stacked ``Residual Units" and each of these units can be formulated as follows:

\begin{equation}
y_l = h(x_l) + F (x_l , W_l ), ~x_{l+1} = f ( y_l ) 
\label{eq:1}
\end{equation}

Where $x_l$ and $x_{l+1}$ are input and output of the $l$-th unit and $F$ is the residual function. The main idea in ResNets is to learn the additive residual function $F$ with respect to $h(x_l)$ with a choice of using identity mapping $ h ( x_l ) = x_l $ \cite{c22}. Moreover, Inception layer is combined with residual unit and it shows that the resulting architecture accelerates the training of Inception networks significantly \cite{c23}.

Recently, due to the limited number of examples and unique characteristics within training samples, there have been interests on one-shot learning methods~\cite{koch2015siamese,santoro2016one,vinyals2016matching}. In the field of Facial Expression Recognition, \cite{jiang2016robust} proposed a sunglasses recovery scheme based on Canny Edge Detection and histogram matching for automatic facial expression recognition by randomly selecting one-shot images from a full database containing images with or without manually-added sunglasses. In \cite{cruz2014one}, a matching score calculation for facial expressions is proposed  by matching a face video to references of emotion.

In the task of FER, in order to achieve better recognition rates for sequence labeling, it is essential to extract the relation of consecutive frames in a sequence. DNNs however, are not able to extract these temporal relations effectively. Therefore a module is needed to overcome this problem. 
Several methods have been proposed for facial expression recognition
in image sequences in order to capture temporal information of facial
movements. For modeling these temporal information, Hidden Markov
Models have been used frequently \cite{c24,c25}. Cohen \emph{et al.}
\cite{c24} proposed a HMM multi-level classifier which combines
temporal information and applies segmentation on video sequences. They
also used a Bayesian classifier for classifying still images. In other
works Dynamic Bayesian Networks (DBN) have been used for facial
expression recognition. Zhang and Ji \cite{c26} used DBN with a
multi-sensory information fusion strategy.

In \cite{c30} it is shown that CRFs are more effective in recognizing
some of human activities like walking, running, etc. Their results show
that CRFs outperform HMMs and even provide good results for
distinguishing between subtle motion patterns like normal walk vs.
wander walk. There are several extensions of CRFs like Latent-Dynamic
Conditional Random Fields (LD-CRFs) and Hidden Conditional Random
Fields (HCRFs) \cite{c31} which incorporate hidden states in the CRF
model. 
In \cite{c33}  these models have been used for facial expression
recognition task.

Discriminative models are reliable tools for the task of sequence labeling\cite{c32,c30,c33}. These models have shown better results in
various modeling problems. Conditional Random Fields are commonly used
in Natural Language Processing tasks like part-of-speech tagging and
word recognition \cite{c8}. Researchers have used these models in
the field of computer vision as well \cite{c33,c30}.


\section{PROPOSED METHOD}\label{sec:3}

Utilizing the Inception layer in Deep Neural Networks has shown remarkable improvement in recognition rates \cite{c18,c4 }. Also, ResNets have shown considerable results in very deep networks \cite{c6, c22}. It looks  intuitive to use the advantages of these architectures to achieve better recognition rates in the field of FER as well. Inspired by the architecture proposed in \cite{c23}, we propose a two-step recognition model consisting of modified Inception-ResNet layers and Conditional Random Fields to perform the task of sequence labeling.

\subsection{Inception-ResNet}

\begin{figure}[thpb]
      \centering
      \includegraphics[width=88mm]{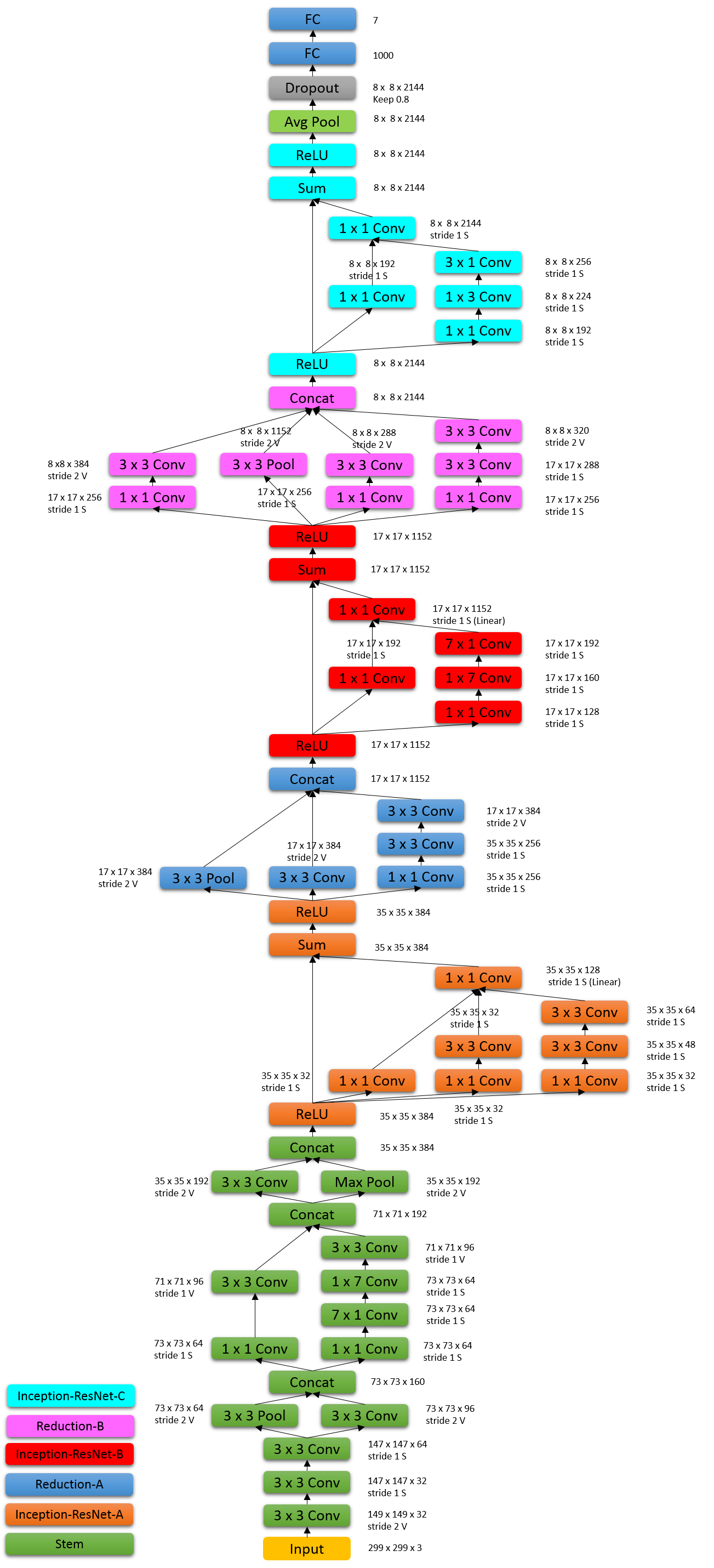}
      \caption{Neural network architecture of our method}
      \label{fig:1}
   \end{figure}

The first part of our proposed model consists of a DNN which is inspired by the Inception-ResNet networks proposed in \cite{c23}. After investigating several variations of Inception layer, we came to the conclusion that Inception-v4 layers achieve better recognition rates comparing to others. Figure \ref{fig:1} shows the DNN part of our proposed module. First, the input images with the size $ 299\times299\times3$ are fed to the ``stem" module. This module is followed by Inception-ResNet-A, Reduction-A (reduces the grid from $35\times35$ to $17\times17$), Inception-ResNet-B, Reduction-B (reduces the grid from $17\times17$ to $8\times8$), Inception-ResNet-C, Average Pooling, Dropout, and two fully-connected layers respectively as it is depicted in Figure \ref{fig:1}. 

All of the convolution layers in the network are followed by a batch normalization layer and all of the batch normalization layers (except the ones which are fed to element-wise summation in Inception-ResNet modules) are followed by a ReLU \cite{c21} activation function to avoid the vanishing gradient problem. The ``V" and ``S" marked layers in Figure \ref{fig:1} represent ``Valid" and ``Same" paddings respectively.

     \begin{figure}[thpb]
      \centering
      \includegraphics[width=70mm]{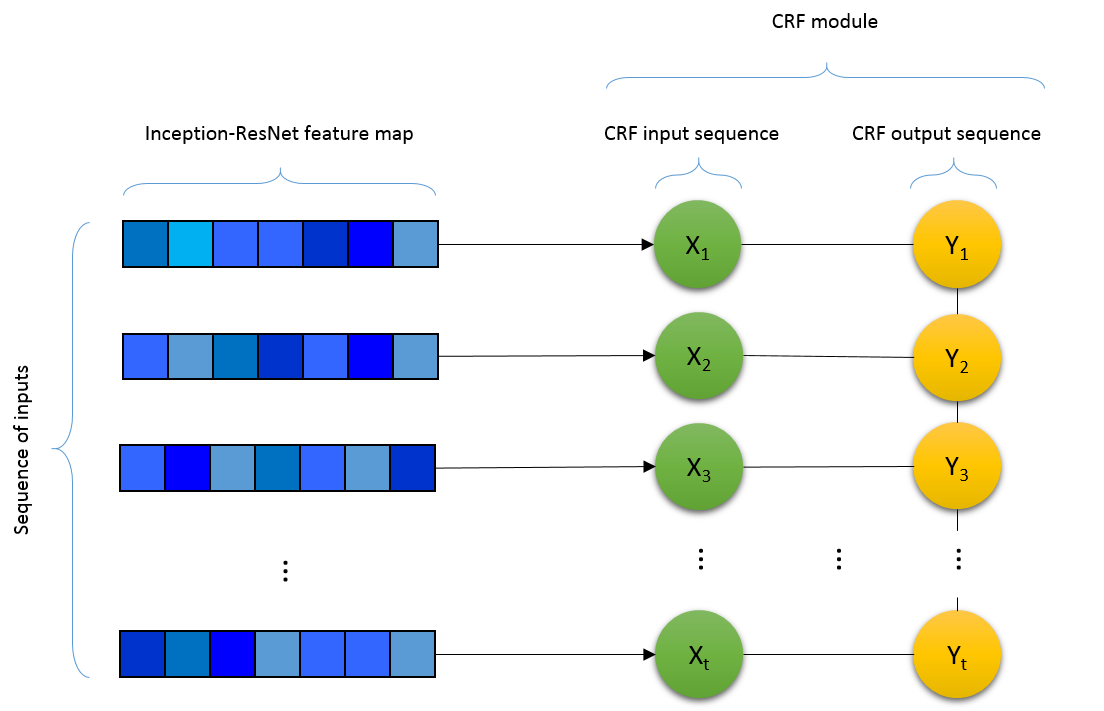}
      \caption{Sequence labeling using CRF }
      \label{fig:CRF}
   \end{figure}

The  proposed  network  was  implemented  using  the Caffe toolbox \cite{c58} on a Tesla K40 GPU. It takes roughly 26 hours to train 200K samples. In the training phase we used asynchronous stochastic gradient descent with momentum of  0.9 and weight decay of 0.0001. The learning rate starts from 0.01 and is divided by 10  every 100K iterations.

\subsection{CRF}

We propose to add a CRF module to the very end of our DNN (connected to the second Fully-connected layer) which will help us to extract the temporal relations between the consecutive frames. As mentioned before, CRFs have shown remarkable results in sequence labeling. Therefore, adding these models to a DNN and use DNN's feature map as input features to CRF will extract the temporal relations of the input sequences. Thus, not only do we extract the spatial relations by our DNN network, we also take temporal relations of these sequences into account using CRFs. Our experiments prove this fact and show remarkable improvement in the sequence labeling task. One of the advantages of CRF over other methods is that CRFs assign the most probable sequence of labels given the whole input sequence which is essential for tasks such as Facial Expression Recognition while other methods such as Long-Short Term Memory (LSTM) \cite{c60} units only take the previously seen frames into account for making the decision.

We train CRF parameters using Gradient Descent method. The main difference  here is that DNNs are trained by batches of the training set. However, in CRF the model is trained in a one-shot manner by the entire samples in the training set during several iterations. That makes our proposed method to be a two-step network. In other words, after training the DNN section of our network by batches of training set using SGD, we feed the obtained feature map to our CRF model and train the model on the whole training set. 

In CRFs our goal is to label sequences $Y_1$, $Y_2$, \dots, $Y_T$ from our
observation sequence $X_1$, $X_2$, \dots, $X_T$ where $T$ is the number of
frames. Given above definitions, CRF model is formulated as follows:

\begin{equation}
P(Y\mid X ; \theta)= \frac{1}{Z(X,\theta)}\exp\Bigg({\sum_{k}{}\theta_k F_k (Y,X)}\Bigg)
 \label{eq:2}
\end{equation}

where,

\begin{equation}
F_k(Y,X)= \sum_{t=1}^{T}{f_k(Y_{t-1}, Y_t,X,t)}
\end{equation}

\begin{equation}
Z(X,\theta )= \sum_{Y}^{}{\exp\Bigg( \sum_{k} \theta_kF_k(Y,X)\Bigg)}
\end{equation}

$Z(X,\theta)$ is the partition function and  $f_k
(Y_{(t-1)},Y_t,X,t)$ is either a state function $s_k (Y_t,X,t)$ or a
transition function $t_k (Y_{(t-1)},Y_t,X,t)$ \cite{c32}. 
Based on \cite{c8}, we use the following objective function for learning the
weights $\theta \textsuperscript{*}$ in CRF considering
the regularization term that maximizes $ \theta^*=\underset{\theta}{\operatorname{arg\,max}}~\mathcal{L}(\theta)$ :

\begin{equation}
\mathcal{L}(\theta )= \sum_{t=1}^{n} \log P\Big(Y\textsuperscript{(j)} \mid \psi(X\textsuperscript{(j)}), \theta \Big) - \frac{1}{2\sigma ^ 2 }\| \theta\| ^ 2
\label{eq:learning_CRF1}
\end{equation}

where its first term is the conditional log-likelihood of the training data. The second term is the $\log$ of a Gaussian prior with variance $ \sigma ^ 2 $ , i.e., $ P( \theta ) \sim exp( \frac{1}{2\sigma ^2} \| \theta \| ^ 2 ) $ \cite{c32} and $ \psi( x_i)$ is the feature map resulted by our previously trained DNN. 

Using Eq. \ref{eq:learning_CRF1}, the final form of $\mathcal{L}(\theta )$ for $n$ training samples regardless of  the regularization term is as follows:

\begin{equation}
\mathcal{L}(\theta )= \sum_{j=1}^{n}\Bigg[  \log  \frac{1}{Z\Big(\psi (X\textsuperscript{(j)}),\theta\Big)}+ \sum_{k}^{}\theta_{k} F_k\Big(Y\textsuperscript{(j)}, \psi (X\textsuperscript{(j)})\Big) \Bigg]
\label{eq:6}
\end{equation}

We used Broyden-Fletcher-Goldfarb-Shanno (BFGS) for learning parameters of our models. We also observed the suitable regularization term value during training process which helped us enhance the recognition rate in our models. Figure \ref{fig:CRF} shows the CRF module of our proposed method. The resulted feature map from DNN part of our method is gathered together as sequences of inputs for the CRF module in order to perform the sequence labeling task.


\section{EXPERIMENTAL RESULTS}\label{sec:4}

\subsection{Face Databases}

Since our method is proposed for sequential labeling of facial expressions, we are not able to use many other databases which include only independent unrelated images of facial expressions such as MultiPie , SFEW , FER2013. We evaluate our proposed method on three well-known publicly  available  facial  expression  databases which contain sequential frames: MMI \cite{c37}, extended CK+ \cite{c38}, and  GEMEP-FERA \cite{c39}.   In this section, we briefly review the content of these databases.

{\bf MMI:} The  MMI   \cite{c37}  database  includes  more  than  20 subjects of both genders, ranging in age from 19 to 62, with different ethnicities (European, Asian, or South American). In MMI's expression pattern, the subject's face first starts from the neutral state to the apex of one of the six basic facial expressions and then returns to the neutral state again. Subjects were instructed to display 79 series of facial expressions,  six  of  which  are  prototypic  emotions.   We  extracted static frames from each sequence, where it resulted in 11,500 images.

{\bf CK+:} The extended Cohn-Kanade database (CK+) \cite{c38} contains 593 video sequences from 123 subjects ranging from 18 to 30 years old; however, only 327 sequences from 118 subjects have labels. All of the sequences start with the neutral state frame and end at the apex of one of the six basic expressions. CK+ primarily contains frontal face poses only. We manually labeled first few frame of each sequence as neutral expression sequences since they are not labeled originally.

{\bf FERA:} The GEMEP-FERA database \cite{c39} is a subset of the  GEMEP  corpus  used  as  database  for  the  FERA  2011 challenge \cite{c41}. FERA consists of sequences of 10 actors displaying various expressions. There are seven subjects in the training set and six subjects in the test set. Five expressions are included in the database: Anger, Fear, Happiness, Relief and Sadness.

\subsection{Results}

In the pre-processing phase, we first register the faces in the databases using bidirectional  warping facial landmarks detected using Active Appearance Models (AAM) \cite{c11}. The resulted transformed face is considered as the face region. After registering the faces, we resize the images to $299\times299$ pixels. We tested several smaller image sizes. Although the network converged much faster in these cases, the recognition results were not promising. 
We evaluated the accuracy of our proposed method in two different sets of experiments: subject-independent and cross-database evaluation. 

In the subject-independent  experiment,  each database  is  split  into training, test, and validation  sets  in  a  strict  subject  independent manner. In FERA, the  training  and  test  sets  are  defined  in  the  database  release, and the results are evaluated on the provided  test set. In MMI and CK+ databases, we consider 10 percent of the subject sequences as the test set and another 10 percent for the validation set. We report the results using 5-fold cross-validation technique. For each database, we trained our proposed network separately from scratch with aforementioned settings. Table \ref{accuracy_subj_ind}  shows  the  recognition rates achieved on each database in the subject-independent case and compares the results with state-of-the-art methods. In order to compare the effect of CRF module in our network, we also provided the results of our network while the CRF module is removed and replaced by a softmax layer.  

\begin{table}[t]
\caption{Recognition Rates (\%) in Subject-independent Task}
\label{accuracy_subj_ind}
\begin{center}
\begin{tabular}{|c||c||c||c|}
\hline
 
  & {\bf Inception-}& {\bf Inception-}&{\bf  }\\
      & {\bf ResNet }& {\bf ResNet }&{\bf  State-of-the-arts}\\
    & {\bf with CRF }& {\bf without CRF }&{\bf }\\
  
\hline
 {\bf CK+}		&	93.04	 &	85.77 & \parbox[c]{2.5cm}{{\center\raggedright84.1\cite{c42}, 84.4\cite{c45}, 88.5\cite{c46}, 92.0\cite{c47}, 92.4\cite{c48}, 93.6\cite{c44}}}\\
 \hline
 {\bf MMI}		& 	78.68 &	55.83 & \parbox[c]{2.5cm}{{\center\raggedright63.4\cite{c48}, 75.12 \cite{c51}, 74.7\cite{c47}, 79.8\cite{c46}, 86.7\cite{c2}, 78.51\cite{c54}}}\\
 \hline
 {\bf FERA}		& 	66.66	&	49.64	 &56.1\cite{c48}, 55.6\cite{c49} \\
 \hline
\end{tabular}
\end{center}
\end{table}

\begin{table}[t]
\caption{Recognition Rates (\%) in Cross-database Task}
\label{accuracy_cross_database}
\begin{center}
\begin{tabular}{|c||c||c||c|}
\hline
  & {\bf Inception-}& {\bf Inception-}&{\bf  }\\
      & {\bf ResNet }& {\bf ResNet }&{\bf  State-of-the-arts}\\
    & {\bf with CRF }& {\bf without CRF }&{\bf }\\
\hline
 {\bf CK+}		&	73.91	 &	64.81		 & \parbox[c]{2.5cm}{{\center\raggedright47.1\cite{c42}, 56.0\cite{c43}, 61.2\cite{c44}, 64.2\cite{c4}}}\\
 \hline
 {\bf MMI}		& 	68.51	&	52.83	 &\parbox[c]{2.5cm}{{\center\raggedright51.4\cite{c42}, 50.8\cite{c2}, 36.8\cite{c43}, 66.9\cite{c44}, 55.6\cite{c4}}} \\
 \hline
 {\bf FERA}		& 	53.33	&	47.05  &  39.4\cite{c4} \\
 \hline

\end{tabular}
\end{center}
\end{table}

In the cross-database experiment, one database is used for testing the network and the rest  are used to train the network. Because of different settings applied in each database (in terms of lighting, pose, emotions, etc.) the cross-database task is a much more challenging task than the subject-independent one and the recognition rates are considerably lower than the ones in the subject-independent case. Table \ref{accuracy_cross_database} shows the recognition rate achieved on each database in the cross-database case and it also compares the results with other state-of-the-art methods. Like before, in ``Inception-ResNet without CRF" column, the CRF module is replaced with a softmax in our proposed network.  .

We should mention that the stated methods in Table \ref{accuracy_cross_database} have not used exactly the same set of databases as ours in the training phase. The results provided in \cite{c42} are achieved by training models on one of the CK+, MMI, and FEEDTUM databases and tested on the rest. The reported result in  \cite{c2} is the best result using different SVM kernels trained on  CK+  and tested  on MMI database. In \cite{c43} several experiments were performed using four classifiers (SVM, Nearest Mean Classifier, Weighted Template Matching, and K-nearest neighbors). The results reported in Table \ref{accuracy_cross_database} are the result of the experiments trained on MMI and Jaffe databases and tested on CK+ (for CK+ results) and trained on the CK+ database and evaluated on the MMI database (for MMI results). In \cite{c44} a Multiple Kernel Learning algorithm is used and the cross-database experiments are trained on CK+, evaluated on MMI and vice versa.  In \cite{c4} a DNN network is proposed using traditional Inception layer. The networks for cross-database case in this work are tested on one of the CK+, MultiPIE, MMI, DISFA \cite{c59}, FERA, SFEW, and FER2013 while trained on the rest.


\section{DISCUSSION}\label{sec:5}

Tables \ref{accuracy_subj_ind} and \ref{accuracy_cross_database} clearly show that adding CRF to our DNN model improves the recognition result in both subject-independent and cross-database tasks. It shows that CRF can successfully distinguish the subtle motion patterns differences which exist between consecutive frames in a sequence.  This improvement is significant in FERA and MMI in both subject-independent and cross-database tasks. 

However, the results in subject-independent task do not show significant improvement from other state-of-the-art methods. FERA shows improvement compared to other methods while CK+ and MMI have comparable results with the state-of-the-art. The main reason is that our network is too deep for these databases. The number of training sequences in these databases is not large enough to be trained properly in a very deep neural network. Therefore the network faces overfitting problem for these small databases. In the Cross-database task which several databases have been combined together, it can be seen that better results have been achieved by the network compared to the state-of-the-art methods.

Table \ref{accuracy_cross_database} shows that in cross-database case, the recognition rate of our purposed network outperforms the state-of-the-art methods. Our method significantly performs better than other methods in CK+ and FERA databases. Recognition rate on MMI is also better than other existing methods although the gap is narrower in this case. 


\section{CONCLUSION \& FUTURE WORK}\label{sec:6}
In this paper, we presented a network for the task of facial expression recognition in videos. The first part of our network is a Deep Neural Network which utilizes the most recent proposed architectures in this field. We use three modified Inception-ResNet modules in our network which  increase the depth and width of the network. This part helps us to extract the spatial relations within a frame. After training this network, we use its feature map in the second part of the network which is a linear chain Conditional Random Field (CRF) for sequence labeling. This part helps us to extract the temporal relations of the frames in a sequence.

We tested our proposed network in two different cases: Subject-independent and Cross-database. Three well known databases are used to evaluate the network: CK+, MMI, and FERA.  We show that incorporating a CRF module to our network improves the recognition rate in all cases. Our network outperforms the state-of-the-art methods in cross-database tasks and achieves comparable results in the subject-independent task in spite of low number of instances in databases in this task.

In the future, we intend to evaluate the method with different variations of CRFs such as higher order CRFs and HCRFs in hopes of extracting better relations in the sequences. Also, we want to evaluate different residual connections in the network for achieving better feature maps for facial expressions.

\section{ACKNOWLEDGMENTS}

This work is partially supported by the NSF grants IIS-1111568 and CNS-1427872. 
 We gratefully acknowledge the support from NVIDIA Corporation with the donation of the Tesla K40 GPU used for this research.


\bibliography{egbib}

\end{document}